\documentclass{article}

\usepackage[final,nonatbib]{nips_2017_modified}

\usepackage[utf8]{inputenc} 
\usepackage[T1]{fontenc}    
\usepackage{hyperref}       
\usepackage{url}            
\usepackage{booktabs}       
\usepackage{amsfonts}       
\usepackage{nicefrac}       
\usepackage{microtype}      

\usepackage{multirow}
\usepackage{graphicx}

\usepackage[square,sort,comma,numbers]{natbib}

\title{Learning to generate classifiers}

\author{
Nicholas Guttenberg \\
Araya \\
\texttt{ngutten@gmail.com}
\And 
Ryota Kanai \\
Araya \\
\texttt{kanair@araya.org}
}

\begin{document}

\maketitle

\begin{abstract}
We train a network to generate mappings between training sets and classification policies (a 'classifier generator')
by conditioning on the entire training set via an attentional mechanism. The network is directly optimized for test set 
performance on an training set of related tasks, which is then transferred to unseen 'test' tasks. We use this to optimize for performance 
in the low-data and unsupervised learning regimes, and obtain significantly better performance in the 10-50 datapoint regime than 
support vector classifiers, random forests, XGBoost, and k-nearest neighbors on a range of small datasets.
\end{abstract}

\section{Introduction}

There are many potential applications of machine learning in medicine, chemistry, small business, and human interaction where the practical costs of obtaining more data are high enough that there is no choice but to deal with small data. In many of these cases, there are plenty of examples of similar problems, but the details of the specific problem are important and should be captured. Model-free machine learning approaches suffer from overfitting in the small-data regime dependent on the regularization techniques which are applied, whereas inference-based approaches which perform well in the extremely low data regime require a particular hypothesis space or model of the generating process to be specified in advance. A compromise between these two extremes is so-called model-based meta-learning, in which a more general machine learning approach is used to learn a model space that can span a wide array of tasks. This can be framed as the meta-task'conditioned on a training set and its labels, what classification policy should we use to maximize generalization performance on test data?' which we can directly optimize in an end-to-end fashion. Rather than the input being a single data point, for this meta-task an input is considered to be an entire data set (similar to \cite{edwards2016towards} and \cite{zaheer2017deep}), split into training and test subsets. By optimizing the entire pipeline directly for generalization performance in the low-data regime, we can transfer knowledge from an ensemble of weakly similar tasks to obtain high performance on unseen tasks for which we do not have explicit models or hypothesis spaces.

In order to process datasets as inputs to a meta-task, Santoro et al. proposed a general class of model called memory-augmented neural networks\cite{santoro2016one}. In these networks, a Neural Turing Machine\cite{graves2014neural} is used to process incoming samples from the training set one by one. However, this requires inputting the data in sequence, and the order of such input sequences can have an impact on the results\cite{vinyals2015order}. We observe that in order to process the network's memory, it is not strictly necessary that the memory by loaded in this sort of serial fashion, but rather we can consider a memory which is already composed of the data which is to comprise the training set and then use attentional mechanisms to address and update this memory, as per the Transformer network architecture\cite{vaswani2017attention}. As such, we present an attention-based memory-augmented neural network which has guaranteed permutation invariance with respect to the order of presentation of its inputs.

While previous work has focused on the image domain, many of the opportunities for low-data models lie in structured data problems, in which there are some set of features which are potentially meaningful in different ways. In this paper, we focus on evaluating the degree to which this type of memory-augmented network is appropriate for such problems. The difficulty posed by this is that, since each feature may be meaningful in a distinct way, and each dataset will likely have different numbers of features, we cannot use a large corpus of related image data in order to learn a model over these types of datasets. We explore the use of synthetic datasets to bridge this gap. 

In inference-based approaches, a model is specified by hand and generally the corresponding inference procedure is derived analytically or simulated in order to perform inference. Here, by formulating the meta-learning task, we can specify models by example and effectively ask the learned network to discover an appropriate parameterization as well as the inference procedure which should be used in order to estimate those parameters. Even when trained only on weakly related synthetic datasets, we obtain good performance in the 20-100 data point regime compared to standard algorithms\cite{pedregosa2011scikit} such as support vector classifiers (linear and RBF), k-nearest neighbors, random forests, and XGBoost\cite{chen2016xgboost}. As the amount of data increases, the relative accuracy gain narrows, but with this method consistently matching performance up to 100 training points. 

Because the problem is formulated over training set/testing set pairs, we can explicitly target statistical effects we suspect may interfere with the robustness and reliability of continuing classification by putting those statistical effects explicitly into the tasks we use to train the classifier generator. For example, we can impose unbalanced classes, covariate shift, and nonstationarity while training the algorithm in order to find classification policies which will be more robust to those effects (to the extent that the specific form of these issues we impose generalizes to the actual problems we are interested in). We can also make use of ancillary problem knowledge such as the expected training set size or available unlabelled data to learn regularization strategies that are tuned to specific data regimes. 

\section{Related work}

The general problem that this method seeks to solve is closely related to the problems of transfer learning\cite{pan2010survey} and one-shot/few-shot learning. In transfer learning, the goal is to take what a model has learned from other weakly related problems and have it apply to a new problem, often in the form of accelerated training or increased data efficiency, while in one-shot learning the concept of transfer often still applies, but there is a stringent constraint on the amount of extra data or training which the model will receive (e.g. it must achieve sufficient transfer with access only to a single representative set of examples of the new problem). 

\paragraph{Meta-learning}
As our method involves learning the process of generating a classification protocol from data as an overarching task, it is related to other meta-learning approaches which broadly fall under the category of learning to learn. These approaches seek to encapsulate some elements of the overall model generation pipeline as tasks in their own right, which can be optimized via training. The approaches taken in this area can be divided into model-based, metric-based, and optimization-based meta-learning\cite{vinyals2017}.

Our approach belongs to the class of model-based meta-learning methods called memory-augmented neural networks\cite{santoro2016one}. We similarly use a memory containing the training data and maximize test set performance over the distribution of datasets, but rather than construct this memory one sample at a time through a Neural Turing Machine, we make use of a self-attention based architecture to process all of the training set and its internal relationships in parallel, in a strictly permutation-invariant fashion. 

Metric-based meta-learning approaches have also been applied to the general problem of one-shot learning, with most of the results being in the image domain. This is a family of approaches which use a combination of an embedding transformation with interpretation of the embeddings in the context of an imposed metric space in order to achieve one-shot learning of similarity relationships in a shared domain (e.g. visual similarity). These approaches include matching networks\cite{vinyals2016matching}, siamese networks\cite{kumar2016learning}, and FaceNet\cite{schroff2015facenet}. The work ``Towards a Neural Statistician''\cite{edwards2016towards} learns latent spaces which explain the statistics of the input (it extracts summary statistics from domain examples), which naturally produces a space that has good properties for metric-based one-shot learning approaches on datasets. 

In optimization-based meta-learning, the process of gradient descent can be formulated as a recurrently applied differentiable transform which should be optimized to reduce the final loss at the end of a period of training\cite{andrychowicz2016learning}. In MAML\cite{finn2017model}, this is cast as a form of transfer learning problem, and as such the parameters of the model used to solve the particular task are still the network parameters (rather than some latent representation of the task inferred directly from the training set), but for which the model as a whole has been adapted such that only a small number of gradient descent steps will be needed in order to obtain high performance. This sort of optimization-based approach has recently been explicitly applied to the problem of few-shot classification\cite{ren2018meta}.

\paragraph{Transfer learning}
Our method can be taken to be a form of inductive transfer learning (in contrast to domain adaptation approaches), where we use the statistics of the overarching set of problems combined with conditioning information about the particular problem we are currently attempting to solve. In our case, the conditioning information directly takes the form of a set of training data. Other related approaches involve  simultaneously training against multiple potential objectives under some constraints or protocols to prevent overfitting to any single object, then performing a degree of fine-tuning to the particular problem case: for example in Elastic Weight Consolidation\cite{kirkpatrick2017overcoming}. Another technique is to learn to stack models using a higher level structure which controls the connectivity of previously learned single-task solvers as in PathNet\cite{fernando2017pathnet}.

Attentive meta-learning\cite{mishra2017simple} is similar to our approach in that an attention mechanism is used to effectively make use of rich conditioning information which acts to specify the particular sub-task which must be solved in a given context. 
  
\section{Method}

Ideally, the architecture we use to formulate the problem of generating classifiers should possess a number of invariances --- it should be invariant to the order of the data points, invariant to the order of the features, and also flexible enough to handle varying numbers of data points, features, and classes. There are a number of methods which explicitly implement permutation invariant or equivariant computations\cite{ravanbakhsh2016deep}. In general, attentional methods also have this property with respect to the data over which attention is applied\cite{vinyals2015order}. As such, we use attentional blocks based on the ``Attention is All You Need'' architecture\cite{vaswani2017attention} in order to obtain invariance over the ordering of the training set and test set data. 

We must also deal with the fact that different datasets will have different feature counts. We tried using non-negative matrix factorization\cite{lee2001algorithms}, PCA, and random projections to provide fixed-length embeddings for the data, and found that random projections significantly outperformed the alternatives. As such, moving forward we simply use random projections followed by normalization to feature-wise zero mean and unit standard deviation in all our presented results.

In order to handle a variable number of classes, we train the classifier generator on a larger number of classes and then mask the probabilities assigned to impossible classes to zero, normalizing the remainder to one. It is also possible via preparing subsampled datasets to explicitly train this kind of generator to produce an 'unknown class' classification (for example, by randomly de-labelling different subsets of classes in one of the training problems and asking the network to assign points belonging to missing classes to this extra category).

\subsection{Network Structure}

\begin{figure}[h]
  \centering
  \fbox{\rule[-.5cm]{0cm}{4cm} \rule[-.5cm]{0cm}{0cm} \includegraphics[width=14cm]{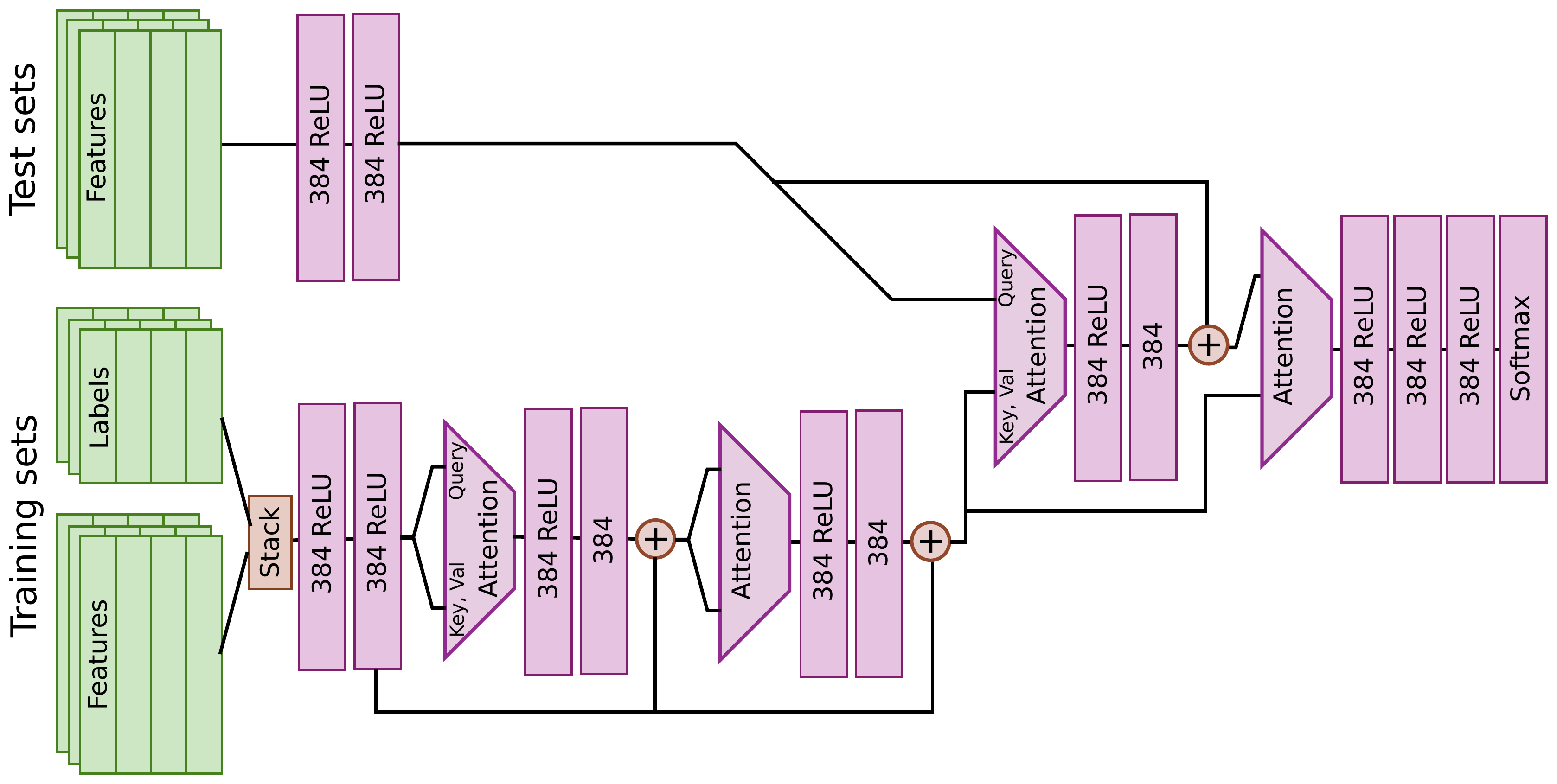}}
  
  \caption{\label{Architecture}Network architecture for the classifier generator.}
\end{figure}

The classifier generator network takes as inputs a set of training points $X_{train}$ which have been randomly projected into a $D$ dimensional space (we present results for networks with $D=4$ and $D=128$) and normalized to have zero mean and unit standard deviation on each axis, a set of one-hot encoded training labels $Y_{train}$, and a set of test points $X_{test}$ projected and normalized according to the same transform as used for $X_{train}$. These are batched, meaning that each of these inputs is a rank-3 tensor. Both training inputs are stacked via concatenation and processed through two initial dense layers with ReLU nonlinearities to increase dimensionality to $384$, obtaining $Z_{train}$. Similarly, the test input is processed through the same structure (but with separate weights) to obtain $Z_{test}$. We find that it is helpful to scale these embeddings by a factor of 10 to accelerate initial training, as it causes there to be more initial variation in the output of the subsequent attentional layers. 

Following the initial embedding, we apply a series of attentional blocks (as per \cite{vaswani2017attention}). An attentional block $Attn(x,y)$ here consists of four attention heads, a query/key space of size $32$, and a value space of size $96$ (such that stacking the results maintains the overall dimension of $384$). The first argument is linearly projected into the key and value spaces, while the second argument is projected into the query space to determine the weighting over $x$. As such, $Attn(x,y)$ can be read as '$y$ searches and returns values from $x$'. We modify the training set embedding via two self-attentional passes $Attn(Z_{train}, Z_{train})$, have the test set attend to the training set over two additional passes $Attn(Z_{train}, Z_{test})$, and then finally apply a series of four dense layers with the final dense layer having a SoftMax nonlinearity with the target number of classes (to be trained against the categorical cross-entropy loss). The overall network structure is shown in Figure \ref{Architecture}.

PyTorch\cite{paszke2017automatic} code for all experiments in this paper is available at \url{https://github.com/arayabrain/ClassifierGenerators}.

\subsection{Synthetic Tasks}

In order to train these networks, we need large datasets of datasets. In this paper, we focus on the use of synthetic datasets to provide this. Normally, synthetic data is a poor test for machine learning performance, but here we are in essence using the synthetic data generator to implicitly specify a hypothesis space which will be learned by the method. Further elaborations on this approach would be to pretrain on synthetic data and then use a smaller set of real related tasks in addition to extensive data augmentation via feature subsampling to fine-tune, but for the purposes of introducing the method we observe that even using just synthetic data for training we can obtain good performance on a variety of real datasets.

The synthetic problem space we use consists of classifying a set comprised of $N_C$ different $N_F$-dimensional Gaussians with random means and covariance matrices. The means are distributed according to a unit Gaussian, and the covariance matrices for each class are each separate random matrices with unit Gaussian components, scaled by an overall problem difficulty factor $\sigma$. While the number of projected input features of the classifier generator is held constant, we vary $N_F$ uniformly between $0.5$ to $3$ times this number during training so that the network will adapt to the type of structure imposed by the random projection operation. We additionally vary $N_C$ uniformly between $2$ and the maximum number of classes for the given classifier generator. We use a statistically uniform balanced set of classes during training, but a given training set is not guaranteed to contain an equal number of examples of each class (and in fact may occasionally contain no members of some classes). 

Training these networks becomes difficult when the number of input features is large, as there is usually a strong initial saddle point associated with the high degree of symmetry of the feature orderings and meanings (e.g. for every problem where a given class is positively correlated with a feature, there will be a problem where it is negatively correlated with that same feature). In order to accelerate the training process, we use a curriculum successively more difficult sets of synthetic problems. This is constructed by doubling the problem difficulty ($\sigma$ range) every time the cross-entropy drops below $0.7$ up to a particular target difficulty level. 

In all cases we use we use Adam\cite{kingma2014adam} with a learning rate $10^{-5}$, and minibatches comprised of 200 datasets. Networks are trained over $1\times 10^{5}$ batches. We refer to two particular models which we compare throughout: $CG_{128,16}^{100}$, which uses $128$ features, $16$ classes, $100$ training points, and a final difficulty $\sigma = 0.1 - 0.3$; and $CG_{32,16}^{100}$, which instead uses $32$ features and a final difficulty $\sigma = 0.2 - 0.6$ but is otherwise the same.


\section{Results}

Here we compare the classifier generator to a number of baseline algorithms on synthetic tasks and real data. In many cases, the dimensions of the problem that the classifier generator is being tested on will be different than the dimensions on which it was trained, and so we use random projections to standardize the problem shape. We found that by ensembling the results of the model over multiple sets of random projections consistently gave significantly better predictions than any single random projection. Additionally, we found that this was true for the baseline algorithms. As such, to standardize the comparison to focus primarily on the differences in the underlying algorithms rather than in the augmentation strategy, in all cases we show results obtained by ensembling over $30$ random projections (to $128$ features, in the case of the baseline classifiers; otherwise, to the input dimension of the classifier generator).

\subsection{Synthetic data}

We will make use of synthetic data generated in the same fashion as the training problems in order to examine the generalization properties of the trained classifier generators to different amounts of data, different numbers of features, different numbers of classes, and different difficulties. For results in this section, we use $20$ randomly sampled problems with $400$ test points each, for each combination of parameter values. We ensure that the training set and test set have at least one member of each class, and compare models based on AUC.

\begin{figure}[h]
  \centering
  \fbox{\rule[-.5cm]{0cm}{4cm} \rule[-.5cm]{0cm}{0cm} \includegraphics[width=14cm]{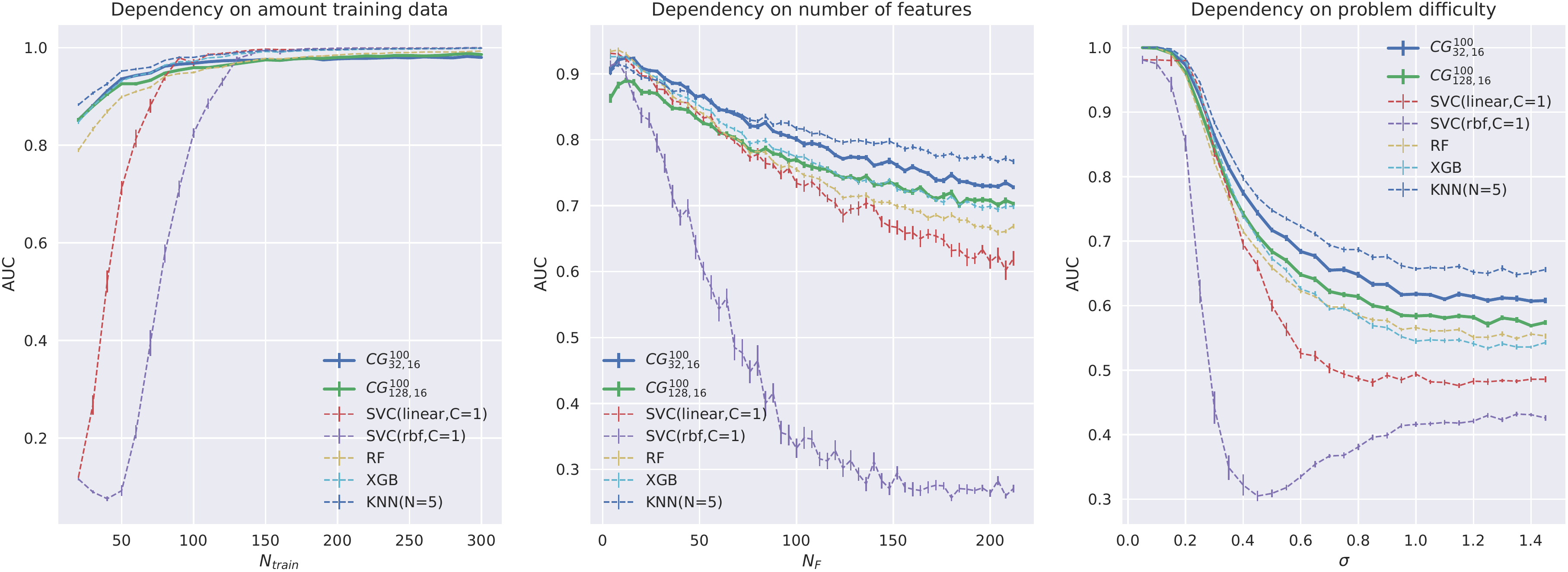}}
  
  \caption{\label{NSweep} Effect of varying amount of training data, relative to the target against which the network was trained.}
\end{figure}

We vary the training set size, feature count, and problem difficulty ($\sigma$) and compare the performance of the $CG_{128,16}^{100}$ and $CG_{32,16}^{100}$ models against Scikit-Learn baselines in Fig.~\ref{NSweep}. Despite the specific classifier generators being tuned to particular numbers of features, data points, or difficulties, their performance is fairly consistent even when those quantities are changed significantly. Furthermore, changing the number of input features for the classifier generator itself has a strong effect regardless of the number of features in the actual testing task, which may indicate that the behavior of ensembling random projections is more important there than any kinds of fine-tuning to the specific dimensionality of the problem.

\begin{figure}[h]
  \centering
  \fbox{\rule[-.5cm]{0cm}{4cm} \rule[-.5cm]{0cm}{0cm} \includegraphics[width=14cm]{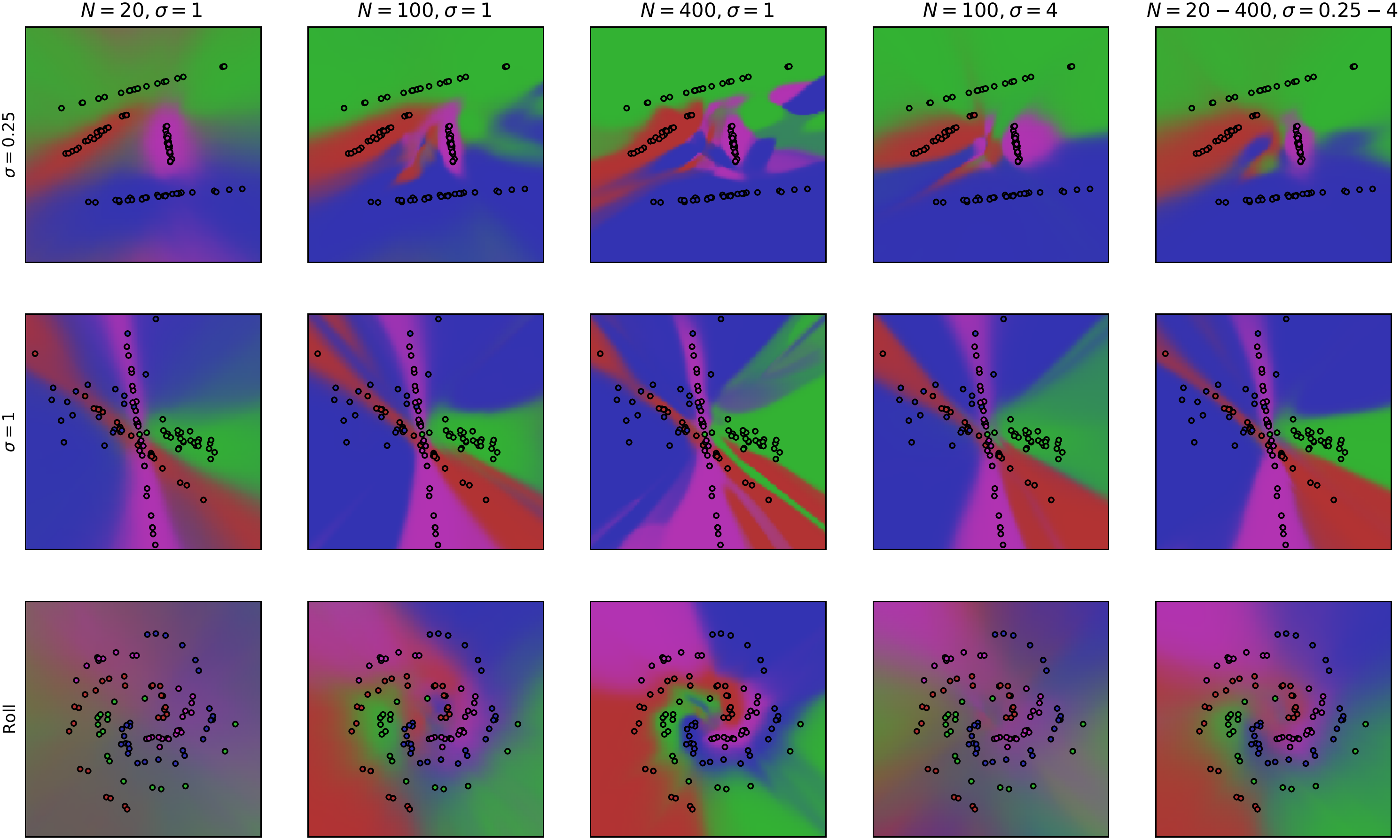}}
  
  \caption{\label{DecisionBoundary}Decision boundaries on problems of different difficulty, with generators trained to expect $20$, $100$, and $400$ points of training data at $\sigma=1$, a generator trained on a harder version of the task ($\sigma=4$), and a model trained over uniformly randomly varied values of both the training data and task difficulty. }
\end{figure}

\paragraph{Visualization of decision boundaries} As there are a number of different factors involved in selecting the synthetic dataset size, train/test split, and difficulty, we would like to obtain an intuitive understanding of how these different considerations influence the type of decision boundaries that the model learns to produce. Since this is difficult in high dimensional spaces, we train a set of classifier generators on 2d tasks with 4 classes. We vary the amount of expected training data, the problem difficulty, and the dispersion of problem difficulty in the synthetic training sets. Each model is trained for $40000$ epochs. Since the 2d problems are relatively easy to learn, we do not need to make use of a staged curriculum of increasingly difficult instances of the problem. In Fig.~\ref{DecisionBoundary} we visualize the decision boundaries for different generators given two synthetic problems from the same space, as well as a swiss-roll dataset to look at how the method may generalize.

\subsection{Small datasets}

We collected a number of small datasets from the UCI repository\cite{Lichman:2013} in order to test the classifier generator on actual data different from its training distribution. We subsample the training data of each dataset down to a variable number $N$ of points, and cross-validate across $100$ train-test splits at each value of $N$. Some datasets supplied an explicit train-test split, but since we subsample the data to control the amount available we only use data from the training set in those cases and report cross validation scores. In all cases, the classifier generator models receive no additional training updates with respect to the particular datasets or any of the other datasets in this section.

In most cases, some data cleaning was necessary in order to standardize the problem. We one-hot encoded categorical values, treating missing values as a unique category. In the case of numerical values, we replaced all instances of missing values with the mean over the dataset. The reported number of features for each dataset ($N_F$) is the final number of columns after one-hot encoding and other preprocessing. We provide the preprocessed datasets with the source code distribution of this paper.

These datasets exhibit some range of class imbalance, with the most extreme case being the cervical cancer dataset which is 94\% composed of a single class (non-detection). Even if some significant features are discovered, in terms of raw accuracy often all classifiers will perform near chance level on these datasets when the amount of data becomes small. However, there are still noticeable differences in AUC score. Results of the classifier generator against the baseline methods are shown in table \ref{UCIResults}. Each point in the table corresponds to $100$ random subsamplings of the dataset (using the same subsamplings for each classifier). 



\begin{table}[t]
  \caption{AUC Results on UCI datasets.}
  \label{UCIResults}
  \centering
  \begin{tabular}{llllllllll}
    \toprule
Dataset & N & $\sigma$ & LSVC & SVC & RF & XGB & KNN & CG & FTCG \\ 
\midrule
\multirow{2}{*}{Immunotherapy\cite{khozeimeh2017expert, khozeimeh2017intralesional}} & 10 & 0.004 & 0.553 & 0.454 & \bf{0.578} & 0.55 & 0.563 & \bf{0.579} & \bf{0.578} \\ 
& 50 & 0.00399 & 0.536 & 0.641 & 0.674 & 0.674 & 0.668 & \bf{0.682} & \bf{0.678} \\ 
\midrule 
\multirow{2}{*}{Forest type\cite{johnson2012using}} & 10 & 0.00274 & 0.681 & 0.351 & 0.94 & 0.922 & 0.887 & \bf{0.948} & 0.943 \\ 
& 50 & 0.000168 & 0.993 & 0.993 & \bf{0.994} & 0.993 & 0.99 & 0.992 & 0.992 \\ 
\midrule 
\multirow{2}{*}{Wine type\cite{forina1990parvus}} & 10 & 0.00241 & 0.863 & 0.733 & 0.984 & 0.981 & 0.965 & \bf{0.99} & \bf{0.989} \\ 
& 50 & 6.56e-05 & 0.996 & \bf{0.999} & 0.998 & \bf{0.999} & 0.997 & 0.998 & 0.998 \\ 
\midrule 
\multirow{2}{*}{Cryotherapy\cite{khozeimeh2017expert, khozeimeh2017intralesional}} & 10 & 0.00416 & 0.834 & 0.661 & \bf{0.898} & 0.886 & 0.851 & 0.89 & 0.891 \\ 
& 50 & 0.00105 & 0.953 & 0.955 & \bf{0.967} & \bf{0.968} & 0.943 & 0.965 & 0.965 \\ 
\midrule 
\multirow{2}{*}{Chronic kidney\cite{chronickidney}} & 10 & 0.00367 & 0.885 & 0.889 & 0.951 & 0.937 & 0.927 & \bf{0.968} & \bf{0.969} \\ 
& 50 & 0.000246 & 0.994 & 0.98 & 0.995 & \bf{0.995} & 0.971 & 0.988 & 0.987 \\ 
\midrule 
\multirow{2}{*}{Wine quality (red)\cite{cortez2009modeling}} & 10 & 0.00156 & 0.407 & 0.366 & \bf{0.637} & 0.593 & 0.62 & 0.601 & 0.602 \\ 
& 50 & 0.00145 & 0.663 & 0.556 & \bf{0.721} & 0.7 & 0.643 & 0.652 & 0.668 \\ 
\midrule 
\multirow{2}{*}{Echocardiogram\cite{echocardiogram}} & 10 & 0.00426 & 0.578 & 0.453 & \bf{0.681} & 0.62 & 0.659 & 0.655 & 0.658 \\ 
& 50 & 0.00403 & 0.579 & \bf{0.75} & 0.716 & 0.688 & 0.647 & 0.723 & 0.733 \\ 
\midrule 
\multirow{2}{*}{Haberman\cite{haberman1976generalized}} & 10 & 0.00345 & 0.542 & 0.46 & \bf{0.592} & 0.576 & 0.568 & 0.586 & 0.586 \\ 
& 50 & 0.00255 & 0.573 & 0.584 & 0.645 & 0.642 & 0.616 & 0.645 & \bf{0.648} \\ 
\midrule 
\multirow{2}{*}{Iris\cite{fisher1936use}} & 10 & 0.0025 & 0.917 & 0.757 & \bf{0.967} & 0.927 & 0.883 & 0.931 & 0.938 \\ 
& 50 & 0.0003 & 0.995 & 0.993 & \bf{0.996} & 0.993 & 0.985 & 0.966 & 0.974 \\ 
\midrule 
\multirow{2}{*}{HCC Survival\cite{santos2015new}} & 10 & 0.00294 & 0.551 & 0.376 & \bf{0.65} & 0.625 & 0.625 & 0.643 & 0.643 \\ 
& 50 & 0.00152 & 0.714 & 0.763 & 0.778 & 0.78 & 0.738 & \bf{0.789} & 0.787 \\ 
\midrule 
\multirow{2}{*}{Wine quality (white)\cite{cortez2009modeling}} & 10 & 0.00134 & 0.427 & 0.395 & \bf{0.592} & 0.562 & 0.58 & 0.572 & 0.565 \\ 
& 50 & 0.00133 & 0.619 & 0.494 & \bf{0.656} & 0.654 & 0.61 & 0.607 & 0.603 \\ 
\midrule 
\multirow{2}{*}{Horse Colic\cite{horsecolic}} & 10 & 0.00177 & 0.507 & 0.446 & 0.548 & 0.551 & 0.544 & \bf{0.553} & 0.551 \\ 
& 50 & 0.00157 & 0.639 & 0.639 & 0.708 & 0.712 & 0.685 & 0.711 & \bf{0.718} \\ 
\midrule 
\multirow{1}{*}{Lung cancer\cite{hong1991optimal}} & 10 & 0.00347 & 0.54 & 0.427 & 0.592 & 0.575 & 0.579 & 0.588 & \bf{0.598} \\ 
\midrule 
\multirow{2}{*}{Hepatitis\cite{hepatitis}} & 10 & 0.0032 & 0.538 & 0.437 & \bf{0.611} & 0.594 & 0.607 & 0.605 & 0.606 \\ 
& 50 & 0.00197 & 0.528 & 0.656 & 0.68 & 0.659 & 0.649 & \bf{0.691} & \bf{0.69} \\ 
\midrule 
\multirow{2}{*}{Dermatology} & 10 & 0.00245 & 0.391 & 0.138 & 0.944 & 0.907 & 0.901 & \bf{0.961} & \bf{0.959} \\ 
& 50 & 0.000456 & 0.992 & 0.965 & 0.995 & 0.995 & 0.993 & \bf{0.996} & \bf{0.996} \\ 
\midrule 
\multirow{2}{*}{Blood transfusion\cite{yeh2009knowledge}} & 10 & 0.00413 & 0.567 & 0.469 & 0.625 & 0.608 & 0.607 & \bf{0.63} & \bf{0.63} \\ 
& 50 & 0.00263 & 0.648 & 0.581 & 0.673 & 0.676 & 0.665 & \bf{0.704} & \bf{0.704} \\ 
\midrule 
\multirow{2}{*}{Autism\cite{thabtah2017autism}} & 10 & 0.0032 & 0.643 & 0.426 & 0.653 & 0.646 & 0.637 & \bf{0.66} & \bf{0.658} \\ 
& 50 & 0.00162 & 0.869 & 0.834 & 0.852 & \bf{0.878} & 0.793 & 0.854 & 0.856 \\ 
\midrule 
\multirow{2}{*}{Cervical cancer\cite{fernandes2017transfer}} & 10 & 0.00211 & 0.587 & 0.424 & \bf{0.618} & 0.516 & 0.531 & 0.565 & 0.572 \\ 
& 50 & 0.00188 & 0.554 & 0.507 & \bf{0.652} & 0.619 & 0.571 & 0.597 & 0.606 \\ 
\midrule 
\midrule 
\multirow{2}{*}{Average} & 10 & 0.000516 & 0.612 & 0.481 & \bf{0.726} & 0.699 & 0.696 & 0.718 & 0.719 \\ 
 & 50 & 0.000286 & 0.755 & 0.758 & \bf{0.806} & 0.801 & 0.774 & 0.798 & 0.8 \\ 
\bottomrule
  \end{tabular}
\end{table}

The baselines we compare against use Scikit-Learn implementations\cite{pedregosa2011scikit} with default parameters. LSVC indicates LinearSVC with $C=1$, SVC is a radial basis function support vector classifier with $C=1$, RF is a RandomForestClassifier with the default number of estimators, XGB is XGBoost, and KNN is KNeighborsClassifier set to use 5 neighbors. We tried varying the regularization in the case of the linear and RBF models and found that while in a few cases a smaller $C$ value can obtain improvements on these datasets, the lower $C$ values were significantly worse on average, and so we omit those results from the table.

In 12 of the 18 datasets, the classifier generator (either fine-tuned or purely from synthetic data) has a top performing entry at either $N=10$ or $N=50$ or both, with the random forest classifier having a top performing entry on 11 of the datasets, XGBoost on 4, and rbf-SVC on 2. However, the random forest classifier obtained a better AUC on average.

\paragraph{Fine-tuning to real data} In addition to the classifier generators trained purely on synthetic data, we examine using the synthetic data as a pretraining step followed by fine-tuning on similar datasets. To do this, we use a leave-one-out cross-validation strategy, taking the same initial classifier generator and fine-tuning to all but the particular dataset we are evaluating on. In addition to the random projection operation, we perform additional data augmentation by randomly permuting the class labels, feature indices, and by subsampling the training data. We train for an additional $20$ batches, where each batch involves examples randomly sampled from the training set of datasets. 

These results are obtained starting from the $CG_{128,16}^{100}$ model, and are labeled as $FTCG_{128,16}^{100}$. Surprisingly, we find very little additional benefit can be extracted from tuning (at least, on this amount of data) as compared to just using the synthetic tasks. The improvements we do see depend on a fairly short fine-tuning period, disappearing when using much more than $50$ batches of additional training. Furthermore, this is despite a significant change in the cross-entropy of the model over this period of time (generally dropping from around $2$ at the start of fine-tuning to $1.3$ at $20$ batches, and as low as $0.7$ for longer training periods. As such, it seems that this type of model would need to be trained on a very large number of real datasets to maintain it's generalization behavior without overfitting in some fashion.

\paragraph{Architecture hyperparameters and overfitting} We tried variations around the basic architecture --- network layer sizes of $256$ and $512$, using four attentional blocks for processing $x$ and $y$ rather than two, and adding a side-path for deriving summary statistics over the entire dataset (similar to \cite{parisotto2017neural}) by pooling over $x$ and stacking with each $y$ representation. In general, we found that final trained cross-entropies on the same problem set could vary from $0.73$ to $0.86$ on the synthetic tasks, with larger networks and more attention blocks systematically corresponding to lower cross-entropies. However, when applying these networks to the small real datasets, we found a performance peak corresponding to the architecture reported in this paper (layer sizes of $384$, with two attention blocks for $x$ and for $y$), with other architectural choices degrading the average AUC for $N=10$ to as low as $0.674$ (will falloffs being more severe for overly large networks than for overly small networks). On the other hand, training a given network longer generally did not decrease the performance on real data. We interpret these observations as detecting a kind of overfitting.

Because the classifier generators are trained against infinite sets of synthetic data, the sense in which they overfit is a bit different from the usual sense of memorizing the training data. Instead, if the family of training tasks as representing a particular precisely-defined hypothesis space (such as the set of Gaussian classification tasks), and as the networks are  given higher capacity, they more closely adopt this prior. However, there is a gap between capturing the broad elements of the hypothesis space (such as nearby points tending to be similar) and treating points as evidence for inference using the full probability distributions of the training tasks. 

\subsection{Computational costs}

Compared to classic machine learning techniques, neural network based techniques tend to be much heavier computationally. Even when there may be some gain in accuracy, in many cases it's a serious consideration as to whether or not the increased computational burden pays for that gain. In the case of the classifier generator model, the initial training required about two days time and 3GB of memory on a single Titan-X GPU. However, once trained, the entire process of fitting a new dataset and rendering classifications operates as a single instance of inference for the neural network, with additional ensembling easily parallelizable in the batch dimension. The result is that, despite the neural network itself being much more complex than the final trained forms of the other baseline classifiers, the actual process of applying it to a new dataset is relatively fast. 


\begin{table}[t]
  \caption{Training and inference time for different classifiers (seconds)}
  \label{TimingResults}
  \centering
  \begin{tabular}{lllllllll}
    \toprule
    $N_{train}$ & $N_{test}$ & LSVC & SVC & RF & XGB & KNN & CG (CPU) & CG (GPU) \\
    \midrule
    100 & 400 & 0.948 & 1.07 & 0.682 & 9.5 & 0.268 & 1.18 & 0.16 \\
    200 & 400 & 2.41 & 2.78 & 0.912 & 17.7 & 0.473 & 1.59 & 0.133 \\
    400 & 400 & 7.97 & 8.43 & 1.45 & 42.9 & 0.928 & 2.94 & 0.167 \\
    100 & 800 & 1.25 & 1.4 & 0.719 & 8.83 & 0.499 & 2.37 & 0.155 \\
    100 & 1600 & 1.88 & 2.08 & 0.805 & 9.02 & 0.899 & 4.43 & 0.232 \\
    100 & 3200 & 3.13 & 3.39 & 0.95 & 9.51 & 1.72 & 8.4 & 0.383 \\
    100 & 6400 & 5.54 & 6.01 & 1.26 & 10.6 & 3.43 & 15.8 & 0.728 \\
    \bottomrule
   \end{tabular}
\end{table}

To compare the practical costs of using the classifier generator, we ran the network in inference mode on CPU and GPU on synthetic problems of different sizes and looked at the time cost per dataset processed when compared to the Scikit-Learn baselines (which all ran on CPU). All problems had 128 features and 16 classes, but varied in the number of training and test data points. These experiments were run on a system with two Intel Xeon E5-2690 v3 CPUs (totaling 48 virtual cores) and a Titan-X GPU (although in practice only the XGBoost benchmark takes advantage of that parallelism and performed optimally at 8 threads). Timings were averaged over 10 datasets each, randomly projecting to 128 features and ensembling 30 times in all cases. The results are in table \ref{TimingResults}.

Despite being costly to train via gradient descent, the fact that the classifier generator model directly 'trains' on new datasets by performing conditional inference rather than via an iterative process means that it manages to perform comparably to the other reference methods which did require training. On the other hand, k-nearest neighbors, which also works directly from data, consistently performed 3-4 times faster. On GPU, the classifier generator was faster across the board for the tested problem sizes, though presumably GPU implementations of the baseline algorithms would also receive comparable speed-ups. The reference methods tend to scale better when the size of the test set is large relative to the size of the training set, suggesting that  at least in terms of pure computational efficiency, this type of method would best fit a use case where retraining on quickly evolving data distributions is a necessity.



\section{Conclusion}

We have demonstrated a method for learning strategies to classify datasets so as to directly target test-time performance. This allowed us to construct classifier generators tuned to specific families of data distributions as well as specific use-cases including very small datasets. The generated classifiers generally performed well, having a top-performing entry on  12 out of the 18 datasets, and having stable performance comparable with other baseline methods in other cases. 

In the larger sense, by framing the overall task of determining a classification policy as correctly predicting test labels conditioned on the training set and test coordinates, we can use gradient descent to discover strategies against statistical difficulties such as limited data, the need for regularization, covariate shift, nonstationarity, and the use of unlabelled data. Rather than having to design a specific pathway for that information to be used, it is only necessary to design a task set which exemplifies the statistical problem in question.

\subsubsection*{Acknowledgements}

We would like to acknowledge Martin Biehl and Nathaniel Virgo for discussion regarding the contents of this manuscript.

\bibliography{classifier_generator.bib}

\end{document}